# GöwFed

## A novel Federated Network Intrusion Detection System

Aitor Belenguer[a,*], Jose A. Pascual[a] and Javier Navaridas[a,*]

[a]*Department of Computer Architecture and Technology, University of the Basque Country UPV/EHU, Manuel Lardizabal 1, 20018, Donostia-San Sebastian, Spain*

ABSTRACT

Network intrusion detection systems are evolving into intelligent systems that perform data analysis while searching for anomalies in their environment. Indeed, the development of deep learning techniques paved the way to build more complex and effective threat detection models. However, training those models may be computationally infeasible in most Edge or IoT devices. Current approaches rely on powerful centralized servers that receive data from all their parties – violating basic privacy constraints and substantially affecting response times and operational costs due to the huge communication overheads. To mitigate these issues, Federated Learning emerged as a promising approach, where different agents collaboratively train a shared model, without exposing training data to others or requiring a compute-intensive centralized infrastructure. This work presents **GöwFed**, a novel network threat detection system that combines the usage of **Gower Dissimilarity matrices** and **Federated averaging**. Different approaches of GöwFed have been developed based on state-of the-art knowledge: (1) a vanilla version; and (2) a version instrumented with an attention mechanism. Furthermore, each variant has been tested using simulation oriented tools provided by TensorFlow Federated framework. In the same way, a centralized analogous development of the Federated systems is carried out to explore their differences in terms of scalability and performance – across a set of designed experiments/scenarios. Overall, GöwFed intends to be the first stepping stone towards the combined usage of Federated Learning and Gower Dissimilarity matrices to detect network threats in industrial-level networks.

## 1. Introduction

In the era of digitization, the amount of data generated and stored has increased exponentially. The current trend of storing and analyzing any digital transaction, combined with the price reduction of storage devices and infrastructures, has caused an outburst of database sizes. In parallel, the number of Internet of Things (IoT) devices is increasing due to the establishment of domestic intelligent gadgets (domotics), the spread of smart cities and the rapid advancement of Industry 4.0, in which has come to be called the Industrial Internet of Things (IIoT). The information generated by those devices is highly appreciated by big data conglomerates, which rely on data analysis for Business Intelligence and understanding market trends with the ultimate goal of improving products and services. As a consequence, data has evolved into a highly valuable asset that needs to be protected.

Cybersecurity has become an essential element in order to avoid data leakages, malicious intrusions, service availability denials and so on. However, the area involving information security is uncertain and needs to be constantly readjusted in line with the emergence of new attack patterns, in a continuous *arms race*. When cybersecurity firstly appeared, the number of computers was insignificant, and they were reserved for professional usage. In those days, ubiquitous, fully-sensorized devices generating huge quantities of network traffic and containing tons of sensitive information neither did exist nor were envisioned. In this context of security preservation, **Intrusion Detection Systems (IDS)** play an important role by monitoring system activity to proactively detect potential attacks. The evolution of threat detection systems has evolved in tandem with the development of new Machine Learning (ML) techniques. The first generation of IDS was rather rudimentary and simply relied on collating system events against manually updated tuples of a signature database. However, these methods were quickly found to have severe limitations, most critically, in terms of flexibility. Primarily, they lacked proactivity in the sense that they were unable to detect new threats that were not in the signature database. Secondly, the period from when an attack was first discovered until new signatures were produced and updated in the IDS was potentially lengthy, leaving systems vulnerable for long periods of time.

As a mitigation, second generation IDS started to gradually incorporate some form of intelligence to detect new threats. This way, they were capable of automatically learning attack patterns using basic ML models, e.g., Support Vector Machines (SVM), Random Forests (RF) and so on. The evolution continued with the incorporation of Deep Learning (DL) techniques, which contributed to the advent of more accurate and sophisticated models, e.g., Multilayer Perceptrons (MLP), Recurrent Neural Networks (RNN) and many others.

*Corresponding author

aitor.belenguer@ehu.eus (A. Belenguer); joseantonio.pascual@ehu.eus (J.A. Pascual); javier.navaridas@ehu.eus (J. Navaridas)

ORCID(s): 0000-0003-2079-0137 (A. Belenguer); 0000-0001-5355-6537 (J.A. Pascual); 0000-0001-7272-6597 (J. Navaridas)

As these systems kept improving in terms of accuracy and new threat detection capabilities, the next natural step is to allow devices to share information about newly detected threats so that new attack vectors are promptly recognized by all parties even if they do not have first-hand involvement. This way, the global impact of new attacks can be reduced. One possible way of achieving this is the incorporation of centralized learning, in which different parties contribute to the training of a complex model by sending their local data to a centralized computing infrastructure. The whole training process is typically performed in a data center (cloud) which will then distribute the new model parameters to all involved parties.

Nonetheless, performing centralized learning could be infeasible due to traditional information sharing approaches that deal with data in a raw way. That could cause network traffic flow struggle; especially in cases where low-resource IoT devices are the main communication agents. Moreover, sharing raw data to third parties is generally discouraged and, indeed, could violate regulations involving data management policies[46]. Therefore, using collaborative learning approaches with strict data protection policies is an interesting alternative to achieve good reliability and scalability, as well as a privacy-friendly infrastructure.

In this context, **Federated Learning (FL)** has emerged as a promising tool to deal with the information exchange of different parties and sensitive data exploitation challenges. FL is an *avant-garde* ML technique that has gained special interest in IoT computing for its reduced communication cost and privacy preserving features[6]. Following an on-device policy, raw data located in the end-devices never needs to leave them. Instead, it is used to learn internal models whose local model parameters can be shared. The local parameters from agents are aggregated into a global model following some predefined rules – e.g., by averaging them as in FedAVG[6]. These consolidated global parameters are sent back to each agent and the process is iterated until convergence is achieved. Thus, knowledge acquired by collaborating devices is pooled to improve the overall metrics of each local model and obtain improved training scores. Based upon the evolution of IDS, we are convinced that FL will conform the backbone of new generation IDS. While FL is a relatively recent technique and its application to IDS technologies is still very limited, our proposal intends to fill this gap by introducing a novel FL-IDS which can adapt to attacks detected by one or multiple devices so that all the involved agents are able to detect new network threats.

In particular, we envision a scenario similar to Figure 1 where several agents connected to, for instance, an industrial network are monitoring bypassing traffic and building a federated model which learns the regular behavior of the network and is able to detect possible attacks by noticing traffic anomalies distributively. These agents can be incorporated into regular network components, or added in the form of a specific device instrumented to carry out this function. Initially, upon any detected threat, agents would raise alarms to the network administrators, but ultimately, the objective

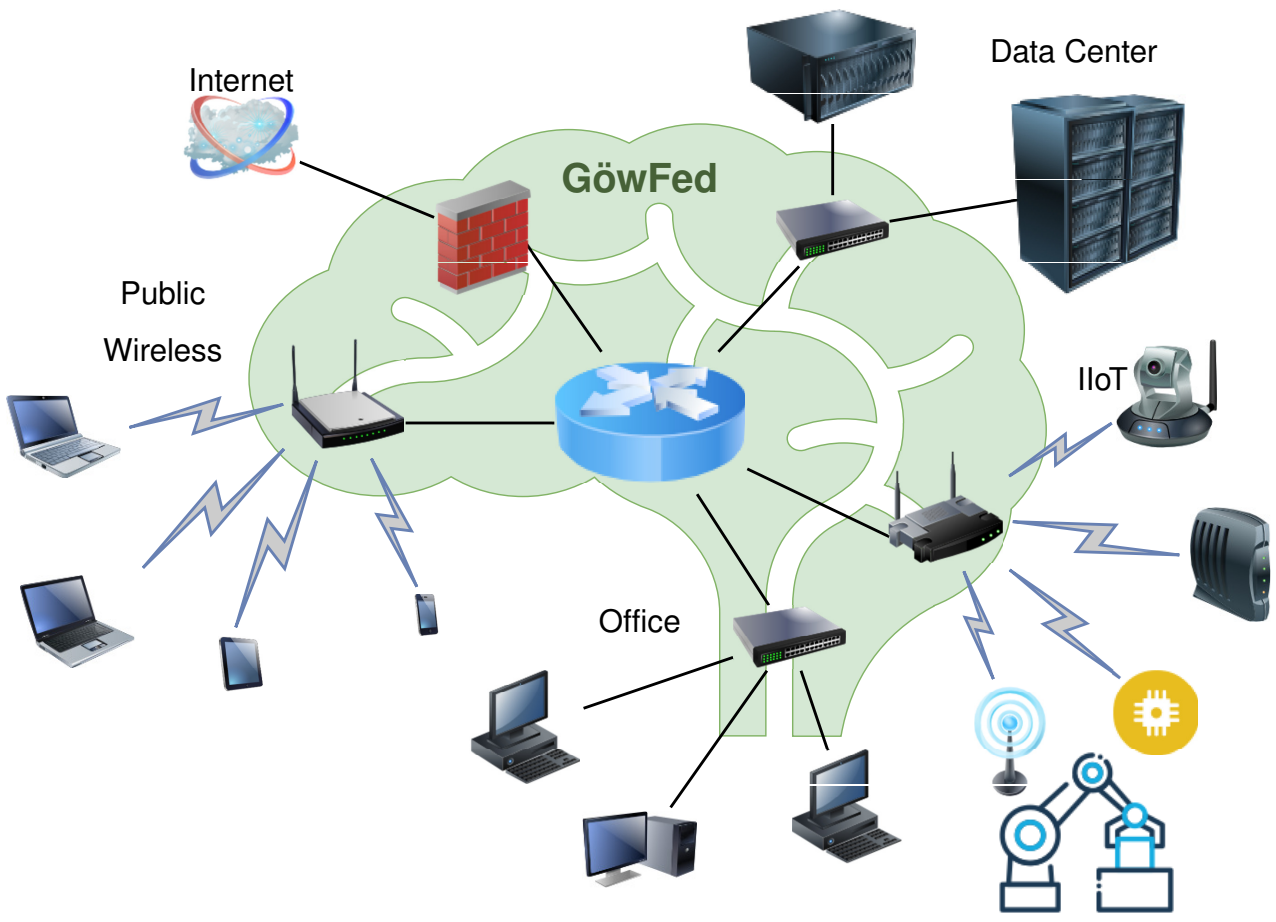

**Figure 1:** Example scenario for a federated intrusion detection system. An advanced manufacturing company relies on GöwFed to protect their industrial, office, data center, public and core networks. Network agents monitor different parts of the network and share knowledge about newly discovered threats.

would be for them to orchestrate defensive measures without human intervention.

Specifically, in this paper we introduce GöwFed, a FL-IDS which leverages Gower distance matrices[15] in combination with state-of-the-art FL techniques (FedAVG, Attention Mechanism (AM)[30], etc.) to streamline the detection process. Similar to B. Li et al.[27] proposal, the agents will monitor network traffic generated by several IoT/IIoT devices, learn a local model based on their behavior and collaboratively build a central model by averaging model parameters with other agents. We started by developing and evaluating centralized models relying on Gower distances. Our first approach leveraged a vanilla MLP model which was able to achieve very high accuracy, precision and recall (up to 99%) which justifies the viability of Gower distances for threat detection purposes. Afterwards, we created a similar FL model to work in (simulated) distributed environments which was able to obtain very good accuracy metrics, although to a lesser extent. Our FL approach was later instrumented to incorporate an attention mechanism, which could be useful to deal with model poisoning attacks and should help to ensure faster convergence. Testing the model with two different attention percentages (20% and 80%) shows a disparity of results. With 20% attention, most of the agents barely contribute to the model and so are unable to perform effective detection, whereas the selected group of agents that contribute the most to the model achieve acceptable performance. With the higher 80% attention, the results are more compelling. In this case, most of the agents contribute to the model so they are able to achieve reasonable results, but the small proportion of the agents which have a smaller contribution are unable to steer the global model to be capable to detect their threats. However, this mechanism has advantages in terms of robustness against model poisoning attacks and facilitating convergence. Overall, we found that the Gower distance is well-suited for detecting network threats and that a FL model can be competitive when compared with a centralized one for this purpose – benefiting from the characteristics of FL and collaborative learning.

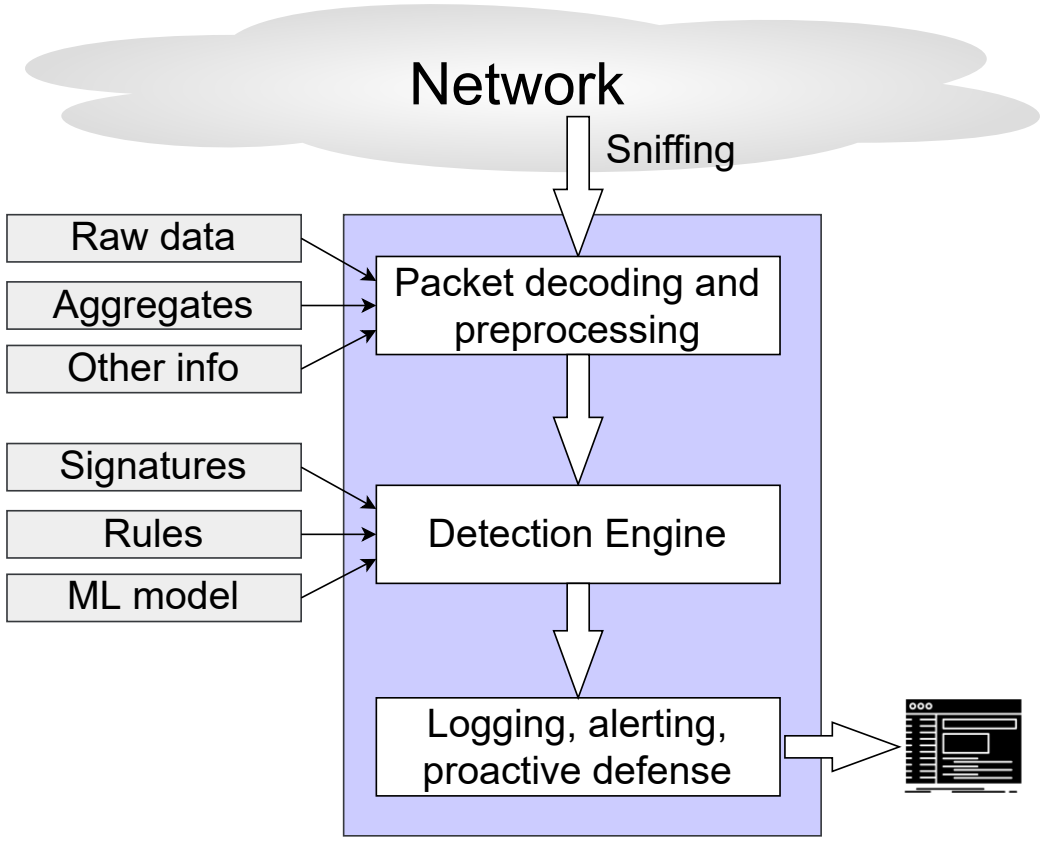

**Figure 2:** Architecture of a NIDS

# 2. State of the art

In order to have a deeper understating of how current FL-IDS technologies work, we point the reader to[5]. Nonetheless, the main concepts are summarized in the following lines.

## 2.1. Network Intrusion Detection Systems

Network Intrusion Detection Systems (NIDS) analyze network traffic with the objective of finding malicious or anomalous patterns that may pose a threat to the devices inside a monitored infrastructure[19]. The typical structure of a NIDS is shown in Fig. 2. It is composed of 4 subsystems: (1) a sniffing system to capture traffic from the network, (2) a pre-processing module to extract the required information from the sniffed traffic either at packet-level, flow-level or other kind of aggregate, (3) a detection engine based on signatures, rules, more complex ML models, or a combination thereof. (4) a relay to send logging information and alerts to the network operators and, if proactive defense is activated, commands to networking equipment.

Recent research[32] has extensively shown that leveraging ML techniques for intrusion detection is a highly successful methodology. This is because these models are capable of learning complex relationships in the data and, in turn, build strong NIDS. The model above can be extended to support federated learning by incorporating a FL engine which interacts with the ML models

### *2.1.1. Datasets for evaluating NIDS*

An important aspect in the life-cycle of advanced NIDS is the evaluation of their detection capabilities. There are many popular datasets available for their evaluation, out of which the most important are gathered together in Table 1. The table uses the following conventions. Raw captures correspond to the availability of the whole captured datagram; usually stored in `pcap` files. Payload features are extracted from the application data in the dataframe and processed using natural language processing, regular expressions or similar techniques. Single Flow Derived Features (SFD) correspond to a collection of packets sharing any property on the IP and transport layers. SFD features are extracted from the aggregation of all packets pertaining to a given flow which is delimited by network events (e.g., the end of a TCP connection, a timeout and so on). Multiple Flow Derived Features (MFD) correspond to the aggregation of information belonging to multiple flow records, containing higher level statistics (e.g., time window delimited flows, last $n$ flows and so on). Finally, dataset labeling could be done manually (M) by a skillful professional; automatically (A) using a rule repository and a script; or on a scheduled (S) way, launching specific attacks in pre-established time windows. It is also possible to merge some of the mentioned methods (MS, AS).

## 2.2. Federated Learning

With the aim of achieving a greater understanding about state-of-the-art FL-IDS, this section provides useful background information on FL. Although FL is the main focus of this project, we also address three additional learning paradigms which are typically used as baselines for benchmark comparisons within IoT/Edge infrastructures.

1. **Self learning (SL)** Neither data nor parameters leave the device; training is performed individually by edge devices. SL can be used as a baseline to measure individual learning ability when no information is shared.
2. **Centralized learning (CNL)** Data is sent from different parties to a centralized computing infrastructure, which is in charge of performing the training with all the received data. CNL is used as a yardstick of the learning ability when the models are built using all available data.
3. **Collaborative learning (CL)** Wraps up custom variants of distributed learning (including FL) where involved agents benefit from training a model jointly. Paul Vanhaesebrouck et al.[45] presented a fully decentralized collaborative learning system, where the locally learned parameters are spread and averaged without being under the orchestration of a centralized authority – in a P2P network.

By definition, FL enables multiple parties to jointly train an ML model without exchanging local data. It involves distributed systems, ML and privacy research areas[22,28], and, since the pioneer FedAVG[6] approach, many new Federated Learning Systems (FLS) have emerged. A general taxonomy describing the difference of those FLS is presented in[28] and replicated in Figure 3 for convenience. This classification is multidimensional and includes the most important aspects of FL architectures including data partitioning, learning model, privacy mechanism, communication architecture, scale of federation and motivation of federation.

**Table 1**
A summary of public datasets available for the evaluation of IDS.

| Dataset | Raw captures | Payload features | Single flow features | Multi flow features | Labeling method [1] | Domain | Year of capture |
|---|---|---|---|---|---|---|---|
| AWID3 [8] | ✓ | ✓ | | | M | wireless IoT | 2020 |
| IOT-23 [42] | ✓ | ✓ | ✓ | ✓ | A | IoT | 2020 |
| TON_IOT [3] | ✓ | | ✓ | ✓ | A | IoT/IIoT | 2019 |
| CICIDS-2017 [37] | ✓ | | ✓ | | S | application traffic | 2017 |
| ISCXTor2016 [17] | ✓ | | ✓ | | S | application traffic (raw/Tor) | 2016 |
| ISCXVPN2016 [16] | ✓ | | ✓ | | S | application traffic (raw/VPN) | 2016 |
| WSN-DS [2] | | | ✓ | | S | wireless IoT | 2016 |
| AWID2 [25] | ✓ | | ✓ | | M | wireless IoT | 2015 |
| SEA [13] | | ✓ | | | A | UNIX commands | 2001 |
| NSL-KDD [41] | | ✓ | ✓ | ✓ | MS | networking | 1998 |

[1] M: manually, A: automatically, S: scheduled.

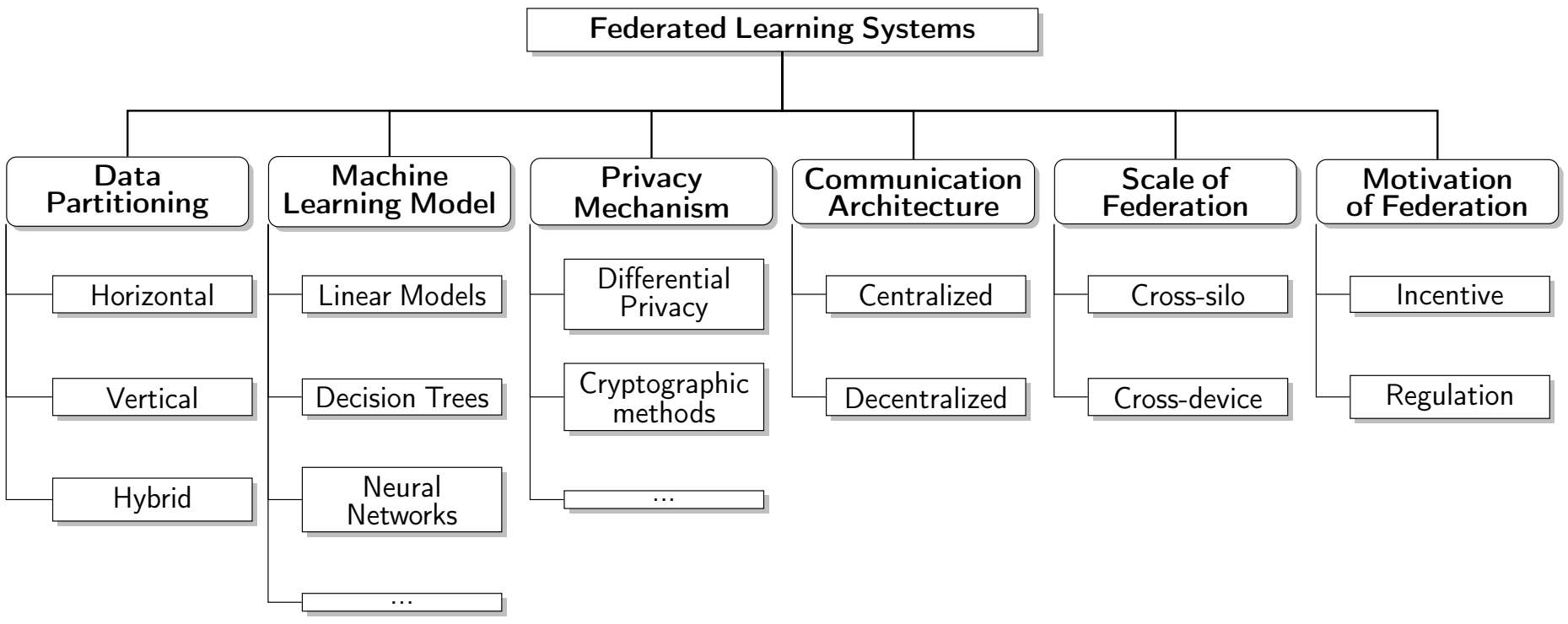

**Figure 3:** A taxonomy of Federated Learning systems (from [28]).

## 2.3. Employing FL in IDS

The combination of both previously explained technologies has become a hot topic of research. Considering that the overwhelming majority of IDS rely on DL models, we introduce a taxonomy based on the DL variants employed by the FL-IDS literature, as illustrated in Figure 4.

The proposed classification is performed taking into account the DL model architecture used on each agent. Since vanilla FL is the *de facto* implementation choice and that many disjoint custom variants exist, it is not possible to perform a tree structure taxonomy by type of FL algorithm used. Hence, existing FL-IDS are split into two major groups depending on the NN architecture; Recurrent Neural Networks (RNN)[31] and Multilayer Perceptrons (MLP)[34]. Each group is respectively divided into two subgroups. The RNN models are further divided based on the neuron architecture into Long Short-Term Memory (LSTM)[20] and Gated Recurrent Units (GRU)[12]. In contrast, MLP models are divided by the model architecture. In particular, models using Autoencoders (AE)[4] are considered an important subclass because they are widely employed in the literature.

The reason for AEs being so common within FL-IDS architecture[7,11,38,39,44] is that their input reconstruction abilities suit themselves very nicely for intrusion detection based on anomaly detection. Once the usual network traffic patterns are learned, anomalies are translated into high reconstruction loss instances. Qin et al.[39] face the challenge of using high dimensional time series with resource limited IoT devices. A greedy feature selection algorithm is employed to deal with data dimensionality issues, as well as a sequential implementation of batch learning is applied to an autoencoder. Tian et al.[44] propose a Delay Compensated Adam (DC-Adam) approach[24] to overcome gradient delay which can lead to inconsistency issues in the learning process. Combined with a pre-shared data training strategy to avoid model divergence in non-IID data scenarios.

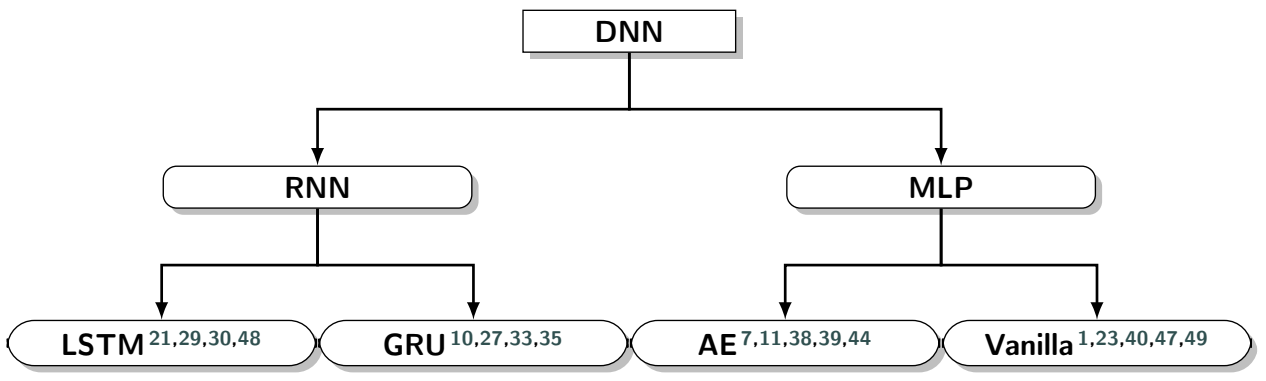

**Figure 4:** Existing Federated Intrusion Detection Systems categorized by model architecture.

The utilization of LSTM NNs is interesting due to their ability to process data sequences which include temporal information[21,29,30,48]. Network traffic flow is considered as a time series, which makes it suitable for a NN using LSTM neurons. Similarly, GRU architectures are a good candidate for processing time series[10,27,33,35]. In contrast to LSTMs, they do not contain an internal memory. However, their simpler architecture makes the learning process lighter which, in turn, renders it suitable for low resource IoT scenarios. DeepFed[27] is an FL-IDS that introduces the concept of Industrial Agents as network monitoring devices and the usage of advanced privacy mechanisms based on Paillier cryptosystem[36] during the FedAVG learning rounds. Dïot[35] also presents relevant advances by identifying device features connected to a local monitoring agent and maintaining a type specific global anomaly repository via FL.

Finally, FL-IDS using custom MLP NN architectures are presented in[1,23,40,47,49]. Beyond the mentioned architectures, those approaches focus on data preprocessing and custom learning variants. Al-Marri et al.[1] merge the advantages of FL and mimic learning[43] by training a teacher (private) and a student (public) model per device to then apply FedAVG and create an IDS. Weinger et al.[47] show how SMOTE[9] and ADASYN[18] data augmentation techniques could accelerate model convergence – reducing communication rounds among agents.

In this paper, we go beyond the state-of-the-art by proposing a model that learns the Gower distances between instances to differentiate between normal and attack traffic. Since the usage of Gower distance does not work with time series as it does not incorporate temporal information, our study will focus on designs based on traditional MLP variants. In particular, we will use a vanilla model to carry out threat detection. This model will be extended to work within a federated architecture and finally, it will be instrumented to use an attention mechanism to provide the capability of dealing with adversarial agents.

## 3. Development

GöwFed'is developed using TensorFlow Federated. Although it is not as complete as FedML in terms of dedicated IoT functionalities, it is extensively documented and has powerful simulation-oriented tools. In this work, we selected the TON_IOT[3] dataset to work in simulated environments. The main reason for choosing this dataset is its versatility; wrapping up vanilla network traffic, modbus devices traffic, raw data captures, well documented datasets as well as its wide usage by state-of-the-art systems. However, designed FL-IDS should behave in a generic way and be extrapolated to other anomaly detection environments – equally working with the other datasets in Table 1 and ultimately in a real environment.

### 3.1. TON_IOT Dataset

In this specific task of monitoring network traffic, the Network dataset subset of TON_IOT (UNSW-IoT20) is chosen – see A. Alsaedi et al.[3] for a detailed description. The TON_IOT contains both categorical and numerical features, making the process of learning more challenging. In addition, as shown in Table 1, it contains `pcap` raw network captures which allow to fully customize input data[a]. Specifically, all the features are taken into account except the timestamp (ts). The main reason to discard the timestamp is that no time trace will be used in the following development; model architecture is not recurrent nor works with time series. The class type is string containing a label: 0 (for *normal* traffic) or 1 (for *attack* traffic). Each class represents, respectively, the 65.07% and 34.93% of total instances. For the time being, a binary classification problem to detect the nature of the datagrams is enough to this initial system; making the problem multiclass and performing novelty detection is left for future work.

[a]A fully detailed description of the features is provided at: `https://tiny.cc/ton_iot`

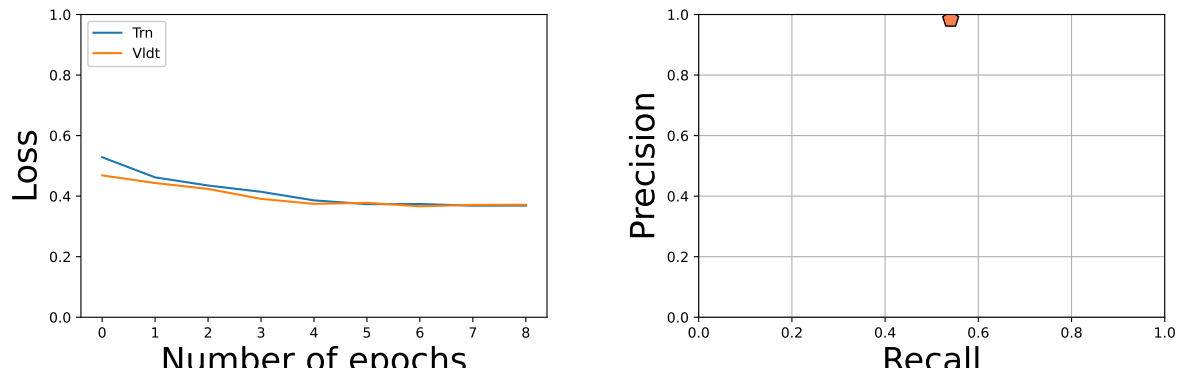

(a) Train and validation loss. (b) Position in the PR space.

**Figure 5:** Results of the **toy classifier** working with exclusively numerical features of TON_IOT dataset.

It is important to note that some data labeled as numerical should be treated as categorical, as they do not encode any magnitude relationship but network information: `src_port`, `dst_port`, `dns_qclass`, `dns_qtype`, `dns_rcode`, `http_trans_depth`, `http_status_code`, `http_user_agent`. However, due to the high amount of different possible combinations, one-hot encoding is not a viable option.

### 3.2. A distance based approach

Working with mixed datasets implies complex data relationships. That is reaffirmed after a few warm up rounds of training a CNL toy classifier[b], where no significant learning is observed – no loss reduction and overall poor performance scores are achieved. Figure 5 shows how this toy classifier working with all the dataset instances[c] is not capable to learn patterns in data: we observe no significant loss reduction and a recall of around 0.5. Therefore, with the aim of addressing the problem from a different perspective, the idea of calculating the Gower Distance among TON_IOT instances emerged.

As mentioned, TON_IOT does not only count with numerical features and transforming categorical data into numerical is not feasible via techniques such as one-hot encoding, due to the number of features increasing excessively. Therefore, working with mixed numerical and categorical data paves the way to use Gower Distance, which we believe should streamline learning data relationships. Gower Dissimilarity (GD) is computed in the following way – being $GD_{ij}$ the Gower Dissimilarity between two instances $i$ and $j$.

$$GD_{ij} = \frac{1}{n} \sum_{f=1}^{n} pd_{ij}^{(f)} \tag{1}$$

Having each instance $n$ different features, either numerical, categorical or mixed. For categorical features, the partial dissimilarity ($pd$) will be 0 if there is a match between the explored couple of features; 1 if there is not. For numerical features, $pd$ is computed by the following partial expressions – the absolute of the subtraction between the specific numerical features divided by the total range of the feature.

[b]Uses the same NN hyperparameters of CNL vanilla version.

[c]Only numerical features due to the unfeasibility reasons mentioned above

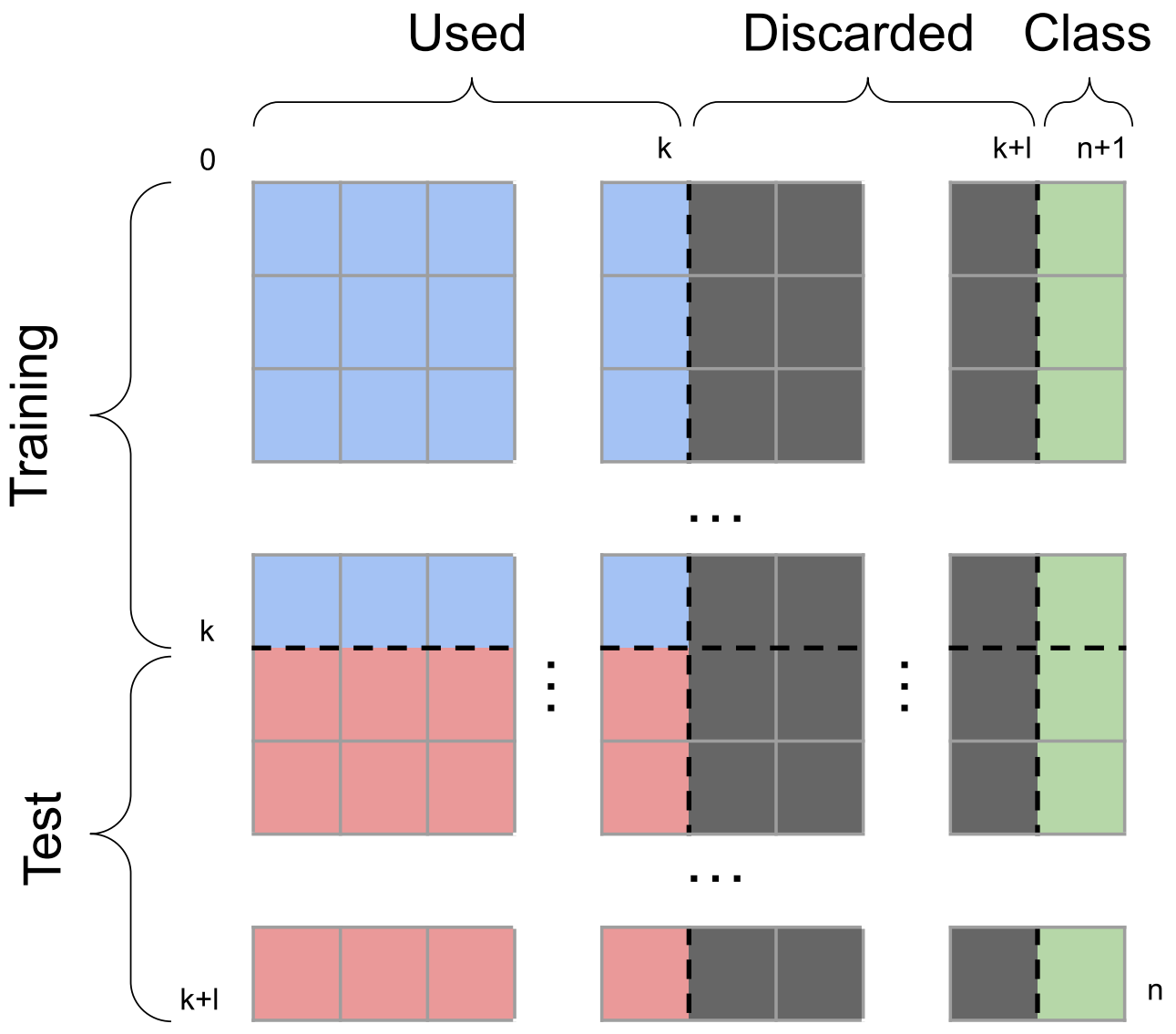

**Figure 6:** Segmentation of generated Gower matrix/matrices to be used in posterior training stages.

$$pd_{ij}^{(f)} = \frac{|x_{if} - x_{jf}|}{R_f} \quad (2)$$

$$R_f = max f - min f \quad (3)$$

After random shuffling the dataset, Gower Distance among instances is computed and sliced, as Figure 6 shows. In both **Gower Centralized (GC)** and **Gower Federated (GF)** or, simply, **GöwFed**, the first $k$ rows and columns (blue) will be used to train the classifier – last $l$ columns (gray) are discarded for containing test information – $k + l = n$. Similarly, the last $l$ instances (red) will be part of the test subset. If an extra validation subset wants to be added, the test partition may be split in an additional subset. At any rate, the computational cost of calculating the initial matrix is $O(k^2)$ but, in practice, once the training matrix is generated, it must not be recalculated. Indeed, it is contemplated that new instances will gradually arrive (data streaming), while network monitoring is performed – adding them to the matrix will have a computation cost of $O(k)$.

Nonetheless, to simulate federated devices, the original dataset is divided into $n_i$ sized disjoint subgroups of instances – an independent Gower Matrix will be computed in each device. Then, $k_i * K$ training rows and $l_i * K$ test instances will be used in the learning process – being delimited by the agent which has the minimum training subset of size $K$ among all.

## 3.3. Designed systems

From this point on, every designed variant will use Gower Distance matrix (or matrices) as input. The designed systems are: (1) a vanilla GC version; (2) a vanilla GF version; (3) a GF version with an AM.

The *Centralized* baseline implementation, GC, is performed using TensorFlow Keras to measure how fast the model can be learned. Model architecture[d] is composed by 6 hidden layers of 128, 64, 64, 32, 32, 2 neurons, respectively, with a dropout rate of 0.15 between hidden layers 4 and 5. The input training data consists of a single Gower Distance matrix as described above.

Our *Federated Learning* version, GF, is implemented using TensorFlow Federated[26], preserving the same model architecture and hyperparameters as the GC version. Finally, the *Attention Mechanism* (AM)[30] approach is implemented using the same NN architecture as GF. In this system, only a pre-established percentage of the agents will contribute to the global model. A proportion, $P$, of the agents with the greatest ROC-AUC values will be selected in each round [e]. Although this mechanism could show generalization issues in scenarios where a small percentage of agents is selected, it can also be interesting to reduce noise and achieve model convergence faster. In other words, selecting the best performing subset of agents can streamline the learning process and improve overall performance. Additionally, this approach is more robust against model poisoning attacks and divergence caused by highly non-IID split information among nodes.

# 4. Implementation details

As mentioned above, creating an open-source, accessible, well-structured code repository[f] is one of the main priorities of this project. Relevant aspects of the implementation are explained in detail below.

## 4.1. Matrix creation

Two modules have been implemented to feed the inputs of the GC and GF versions `create_matrix_cnl` and `create_matrices_fl`, respectively. Additionally, those modules are in charge of balancing the datasets (if required) and distributing the instances among the different FL agents. Finally, separated files containing training and test Gower Matrix partitions will be saved – in the case of the GF module, training and test matrices files will be created per agent.

Gower library is extended by adding custom methods to produce the training/testing partitions, as described in Figure 6. These methods are called `gower_matrix_limit_cols` and `sliced_gower_matrix_limit_cols`, respectively. In the GC version, the existing `gower_matrix` function is called, passing as argument just the training instances, to get the $k \times k$ training matrix – cost $O(k^2)$. Then, *sliced_gower_matrix_limit_cols* is called passing the whole dataset, the number of training instances (to be skipped) and the column number limitation to get the test partition – cost $O(k^2)$.

[d]Fully connected MLPs with a $k$ sized input layer and a 1 sized output layer (*sigmoid*). Using *Adam Optimizer*, *Binary Accuracy* and *Binary Cross-entropy* as training hyperparameters.

[e]Other criteria and their combinations could be used as well, such as picking the nodes with greater F1 score, accuracy and so on.

[f]`https://github.com/AitorB16/GowFed`

On the other hand, in the GF version, the whole dataset is uniformly distributed in small partitions corresponding to each client – the length of the partition with the least number of training instances, $\mathcal{M}$, is stored for future use. This models an IID distribution of the dataset, where the clients are expected to have similar local dataset sizes. Regarding gower library, `gower_matrix_limit_cols` method is called to generate the training matrices – limited by $\mathcal{M}$. In other words, if no data augmentation is performed, the agent with the least training instances will limit future models' input sizes and define $k$. Finally, the `sliced_gower_matrix_limit_cols` method is called to generate the independent test partitions.

### 4.2. Federated Learning

Firstly, the datasets (Gower Matrices) have to be adapted to a `tff.simulation.datasets.ClientData` federated object type in order to be suitable for **TensorFlow Federated** simulation. The federated dataset is represented as a list of client ids, and a function to look up the local dataset for each client id. Although TensorFlow Federated contains its precompiled testing datasets (EMNIST, CIFAR, ...), the `tff.simulation.datasets.TestClientData` class allows creating custom datasets for simulation purposes. However, a series of constraints have to be met before instantiating the class: (1) load the training and test datasets of each client and transform them into independent dictionaries, where the keys correspond to the features and the class of each dataframe; (2) create two global dictionaries, (one for training and the other for testing) where the keys correspond to clients IDs and the values wrap up the dictionaries of step 1. Once the criteria is met, two `TestClientData` instances (training and testing) are created by passing the global dictionaries as arguments – in independent calls.

The backbone implementation of all developed FL algorithms is based on the **SimpleFederatedAveraging** guideline provided by the TFF team – all our custom variants follow the same scheme. A TFF federated algorithm is typically represented as a `tff.templates.IterativeProcess`. This is a class that contains `initialize` and `next` functions. The former is used to instantiate the server, and the latter carries out one communication round of the federated algorithm[g]. The four main components composing the federated algorithms are:

1. **A server-to-client broadcast step:** The server sends the global weights to all the clients taking part in the communication.
2. **A local client update step:** The local gradient is computed on batches of data and then aggregated within the received server weights.
3. **A client-to-server upload step:** The computed local weights are uploaded to the aggregator server.
4. **A server update step:** The server model weights are replaced by the average of clients' model weights (FedAVG); where the importance of each client during the averaging process is proportional to its number of local instances; an intrinsic characteristic of developments based on `SimpleFederatedAveraging`.

In order to manage and customize the orchestration logic of what the server broadcasts to the client and what the client updates to the server, the Federated Core (FC) API is used. This API has three relevant elements to be mentioned:

- Federated data type; a data structure hosted across the clients, where the federated type and the placement are defined (e.g., `float32@CLIENTS` meaning that each client has a `float32` type object).
- `tff.federated_computation`; a specification in an internal platform-independent glue language [14] – functions with well-defined type signatures that can only contain federated operators.
- `tff.tf_computation`; blocks containing TF code without specifying that can be mapped into federated computations via `tff.federated_map` method.

The AM versions maintains basically the same skeleton as GF, but with small variations. In that case, the `next` function of the iterative process is split in two stages. The first one computes the local models of the selected subset of clients and sends the partial results and the computed local weights to the server. Then, the server picks the $k$ best performing nodes according to the best ROC-AUC areas and discards the remaining $n - k$ agents – the $k$ number of nodes varies according to the selection proportion specified in the initialization file. Finally, in the second stage, the selected weights are uploaded to the server and **FedAVG** is performed; modulating the influence on the global model by the number of instances.

## 5. Experimental setup

A series of experiments have been carried out to test GöwFed's (GF) performance and scalability. In order to do so, outputs from a series of analogous CNL versions are explored as the baseline (GC), contrasting their performance against the FL systems. The experimentation has been performed in a machine with the following specs: an Intel i9-7920X CPU at 4.3Ghz with 64GB DDR4 2400MT/s RAM and 2x Nvidia RTX 2080 Ti 11GB GPUs.

### 5.1. Metrics of interest

In our experiments, we collect a number of metrics of interest: accuracy, precision, recall and false positive rate. Out of these, we can compute other important metrics such as the F1 score, the area under the ROC curve and the position in the precision-recall (PR) space. However, for the sake of simplicity we will present in our discussion only the accuracy as a general measure of the detection ability of the different NIDS and the position in the PR space which allows to ascertain, visually, the kind of errors that the detector is producing. A low precision means that the detector is marking a large proportion of the normal traffic as a threat which will result in too many alerts being raised to the network administrators or, even worse, in measures being put in place to deal with these false threats that affect legitimate traffic. In contrast, a low recall means that the detector is not able to recognize attacks so the network is not properly protected. In particular, the PR curve is very informative in unbalanced scenarios, where accuracy alone may produce results that are biased towards the dominating class.

[g] https://tensorflow.google.cn/federated/tutorials/building_your_own_federated_learning_algorithm

**Table 2**
Configuration parameters for the CNL experiments.

| Run name [a,b,c] | Training DS size | Test DS size | Balanced dataset | Epochs | Learning rate | Batch size | Seed |
|---|---|---|---|---|---|---|---|
| GC_SB | 10000 | 2000 | ✓ | 100 | 0.0001 | 64 | 26 |
| GC_SU | 10000 | 2000 | × | 100 | 0.0001 | 64 | 26 |
| GC_MB | 20000 | 4000 | ✓ | 100 | 0.0001 | 64 | 27 |
| GC_MU | 20000 | 4000 | × | 100 | 0.0001 | 64 | 27 |
| GC_LB | 40000 | 8000 | ✓ | 100 | 0.0001 | 64 | 28 |
| GC_LU | 40000 | 8000 | × | 100 | 0.0001 | 64 | 28 |

[a] : GC: Gower Centralized Learning. [b] S: Small; M: Medium; L: Large.
[c] B: Balanced; U: Unbalanced.

**Table 3**
Configuration parameters for the FL experiments.

| Run name [a,b,c] | Node numbr | Training DS size | Test DS size | Balanced dataset | Total rounds | Nodes /round | Local epochs | Srvr lrng rate | Clnt lrng rate | Training batch | Test batch | Seed |
|---|---|---|---|---|---|---|---|---|---|---|---|---|
| GF_SB | 10 | 30000 | 5000 | ✓ | 100 | 4 | 10 | 0.0001 | 0.00001 | 128 | 32 | 26 |
| GF_SU | 10 | 30000 | 5000 | × | 100 | 4 | 10 | 0.0001 | 0.00001 | 128 | 32 | 26 |
| GF_MB | 20 | 60000 | 10000 | ✓ | 100 | 8 | 10 | 0.0001 | 0.00001 | 128 | 32 | 27 |
| GF_MU | 20 | 60000 | 10000 | × | 100 | 8 | 10 | 0.0001 | 0.00001 | 128 | 32 | 27 |
| GF_LB | 40 | 120000 | 20000 | ✓ | 100 | 16 | 10 | 0.0001 | 0.00001 | 128 | 32 | 28 |
| GF_LU | 40 | 120000 | 20000 | × | 100 | 16 | 10 | 0.0001 | 0.00001 | 128 | 32 | 28 |

[a] GF: Gower Federated Learning – GöwFed. [b] S: Small; M: Medium; L: Large. [c] B: Balanced; U: Unbalanced.

## 5.2. Gower Centralized

We use a Gower Centralized model as the baseline to compare our novel GF approaches. Table 2 summarizes the different run configurations. In our experiments, we keep track of training and validation losses. Furthermore, we store a copy of the running configuration, the learned h5 model and a large collection of metrics: accuracy, precision, recall, F1 score and ROC-AUC.

Our experiments are classified in three subgroups depending on the selected training/test subset size; Small (S), Medium (M) and Large (L). As shown in Table 2, each subgroup doubles in size from the previous one, from 10K to 20K to 40K instances. Moreover, two variants, balanced (B) and unbalanced (U) are used. In the balanced experiments, the dataset is pre-processed to obtain the same proportion of 'attack' and 'normal' instances, i.e. 50% of instances each. However, in unbalanced experiments, the proportion of each kind of instance is maintained from the original dataset (65.07% 'normal' and 34.93% 'attack'). Finally, the rest of the parameters are deeply analyzed in Appendix 7.

## 5.3. Gower Federated

A similar set of experiments is used to test GF performance and scalability. Working with Gower Dissimilarity matrices requires a high amount of system memory. That is why, a pseudo-fixed amount of instances is used at each client – obtained by uniformly distributing the dataset as explained in Section 3.3. The local dataset sizes per experiment are shown in Figure 7 and are the same for all GF models. Moreover, the total number of instances and the number of agents taking part in each averaging round, will be proportional to the total number of agents. In other words, the data subset size in each agent will be similar (around 3,000 instances), regardless of the number of agents. Specifically, six configurations, with different seeds, have been tested to measure the scalability of the system, which are summarized in Table 3.

Again, for each experiment we capture a series of attributes and metrics containing each nodes' information: agent id, agent training dataset size, accuracy, precision, recall, F1 score and ROC-AUC area. Those metrics are obtained by evaluating the averaged global model against the test subset matrix of each agent – each agent is expected to converge to the same global model after the `SimpleFederatedAveraging` emulation. The experimentation is performed for all the systems discussed in Section 3.3 – AM version is run twice with different attention parameter (0.2 and 0.8) to simulate tighter attention constraints.

# 6. Results

After running the simulations, a deep analysis of the results has been carried out. First, we discuss the results of the centralized vanilla version. Then, we contrast these with the results obtained with the vanilla, and AM versions of the federated systems.

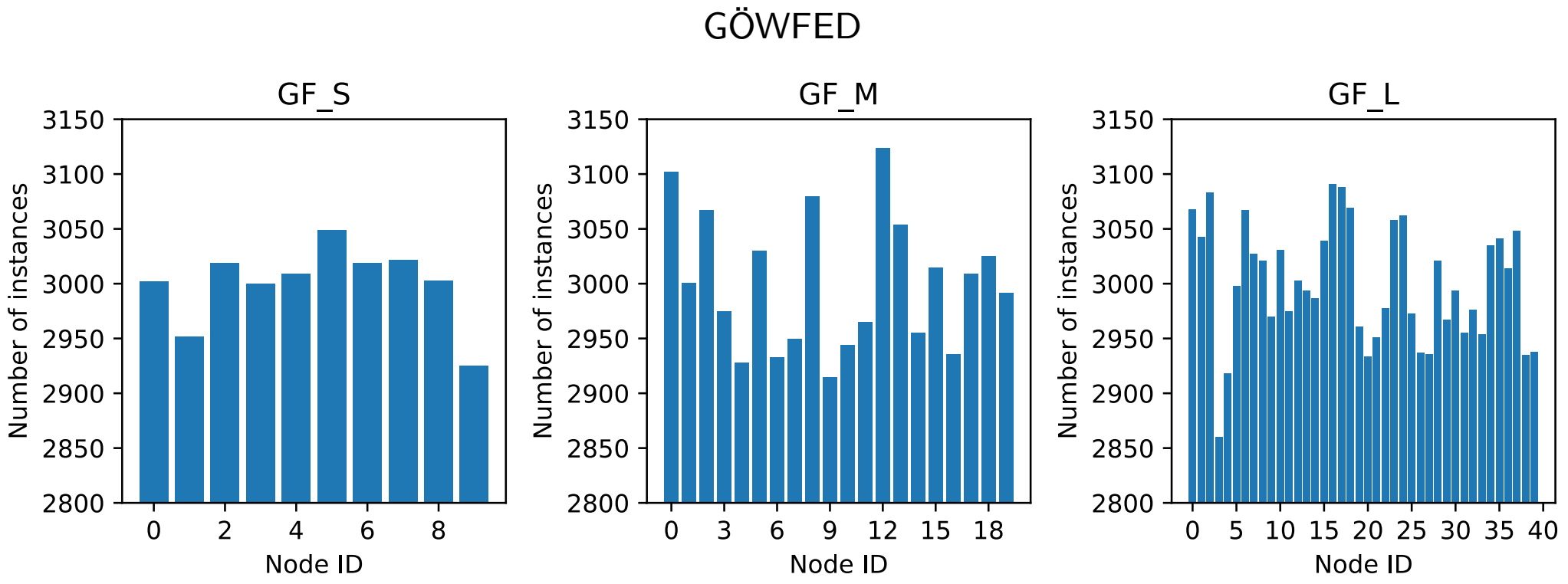

**Figure 7:** Training dataset partition sizes per agent ID in every GF version.

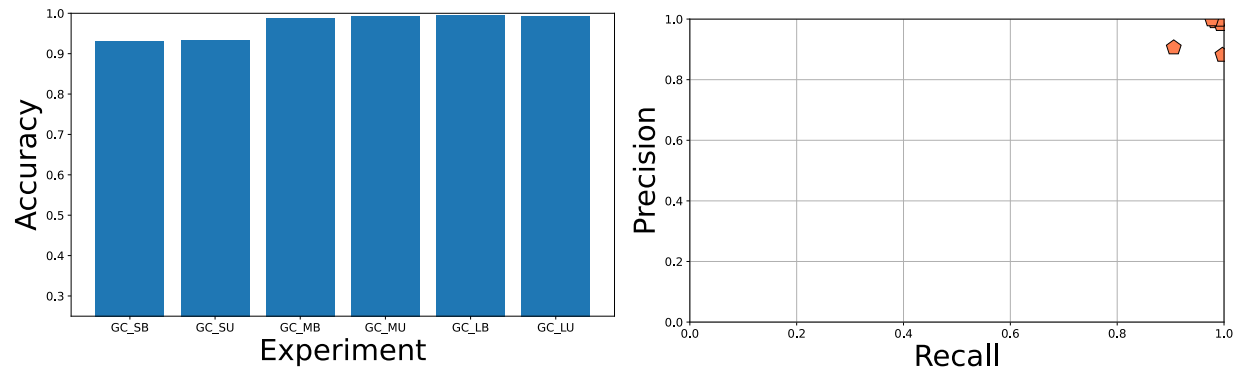

(a) Accuracy. (b) Position in the PR space.

**Figure 8:** Results of learned **vanilla GC** models in the test partition of each experiment.

### 6.1. Gower Centralized

As shown in Figure 8, the results obtained by the GC vanilla version are excellent. PR space shows models trained with the large and medium subsets obtain very high precision and recall (the median point of the experiments is $[0.987, 0.987]$) and the median accuracy is 0.989. The models trained with the small subsets, however, obtain a lower performance. The median accuracy is 0.931 and the median point in the PR space is $[0.893, 0.951]$ – of both balanced and unbalanced experiments. The models, trained with the medium subsets, obtain remarkably improved performance compared to the previous experiments. The median accuracy is 0.989 and the median point in the PR space is $[0.987, 0.988]$. The models trained with the large subsets, as expected, obtain the highest performance. The median accuracy is 0.993 and the median point in the PR space is $[0.997, 0.984]$. For the balanced dataset, a very high recall is maintained, being 0.992 the median. In contrast, with the unbalanced subset, the recall is also reduced, being 0.977 the median.

### 6.2. GöwFed

Figure 9 shows the performance per agent of the shared global model built with the GF vanilla version. In the figure, each bar or point corresponds to a specific agent ID – the results correspond to the trained global model in the test partition of each agent. Regarding the obtained results, the median point of the experiments in the PR space is $[0.888, 0.960]$ and the median accuracy is 0.930 – being comparable to the ones of GC. The models trained with the small subsets, obtain the lowest performance. The median accuracy is 0.829 and the median point in the PR space is $[0.784, 0.925]$ – of both balanced and unbalanced experiments. The models trained with the medium subsets, obtain higher performance compared to the previous experiments. The median accuracy is 0.955 and the median point in the PR space is $[0.916, 0.992]$. Although the models trained with the large subsets do not obtain the highest performance, the difference is insignificant – that could be explained by an increase in convergence complexity. The median accuracy is 0.922 and the median point in the PR space is $[0.880, 0.958]$. Moreover, the PR space results show an overall good performance of the global model in each nodes' test partition.

### 6.3. GöwFed with Attention Mechanism

Finally, regarding the versions which incorporate AM. The strict-attention development, with 0.2 of best performing agents, obtains very dissimilar results among the agents – Figure 10 shows the mentioned disparity. As it is expected, some agents never contribute to the averaged model and a significant proportion of them are not capable to learn all the threat patterns – the learned model does not generalize well, and bad results are achieved. In other words, the gap between the good and bad performing agents gets accentuated as the total number of agents increases. However, with looser attention, using the 0.8 best performing nodes, the results are overall comparable to the vanilla GF version – with an acceptable performance of the global model in the majority of the nodes; Figure 11. Despite not being the current scenario, this mechanism could be interesting to ensure convergence in cases where adversarial/byzantine agents are present. Nevertheless, AM causes decrement of performance in scenarios where data is IID distributed among nodes – as the current one.

### 6.4. Discussion

As could be expected, GC outperforms GF in the vanilla version experiments. However, the comparison is not completely fair due to the number of training rounds that FL versions could require to achieve the same convergence levels than CNL versions. As it is described in Tables 2 and 3, the GC models are trained with 100 epochs and the GF models with 100 communication rounds – 10 local epochs per round are performed in each round. Furthermore, an early stopping criterion is used in the GC systems, with 2 rounds of patience, that causes each experiment to have a variable number of training rounds. GF systems are forced to stop at 100 communication rounds, as a timeout – keep in mind that epochs in FL systems are performed independently by each agent. Figure 12a shows that training and validation losses in each vanilla GC experiment end up converging into the same values. Nonetheless, training and validation losses of vanilla GF experiments do not converge equally well – see Figure 12b.

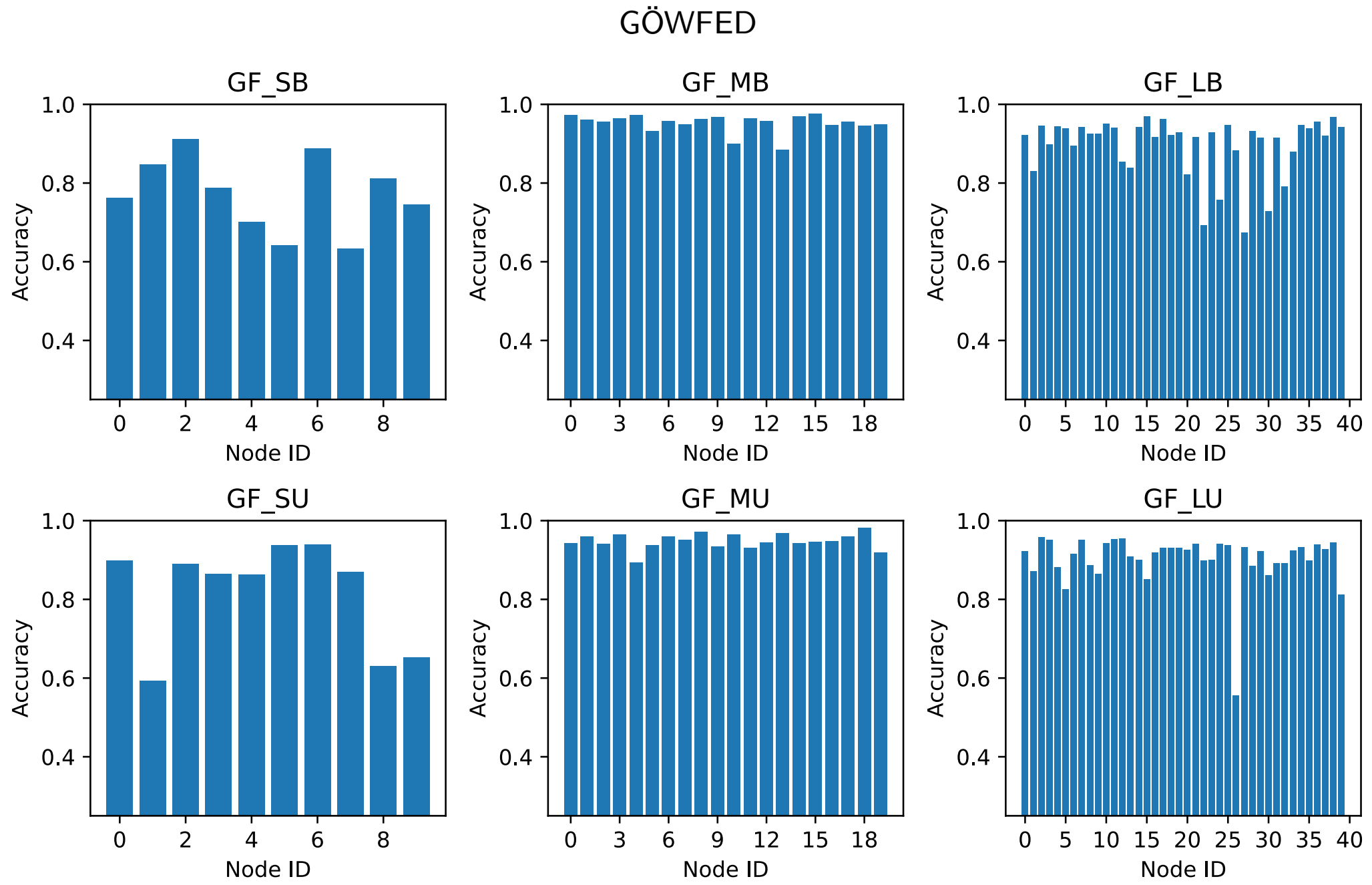

(a) Accuracy per node ID.

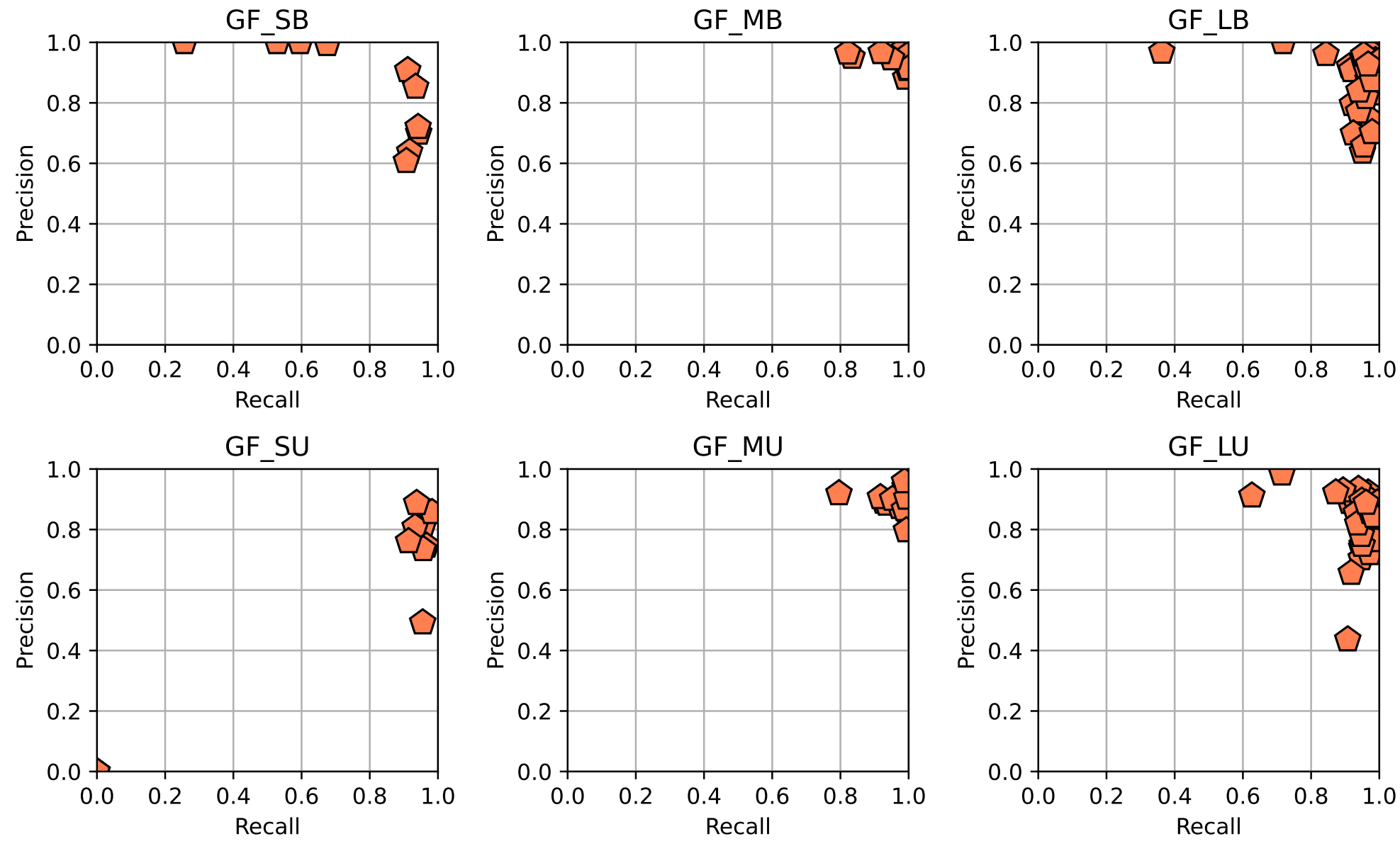

(b) Position in the PR space.

**Figure 9:** Results of learned **vanilla GF** models in the test partition of each agent per experiment.

Moreover, the comparison is not strictly fair due to scalability reasons. In our experimental platform (Section 5), working with CNL Gower Matrices of more than 40,000 instances is computationally infeasible, because their quadratic nature makes them exceed the memory capacity of our system. However, with the federated version, as the instances are spread out through all the agents, we are capable of using a much larger collection of network traffic. For instance, the experiment GF_LU is using a total of 120,000 instances across all agents. This shows that using our Gower Distance approach in distributed systems is much more scalable.

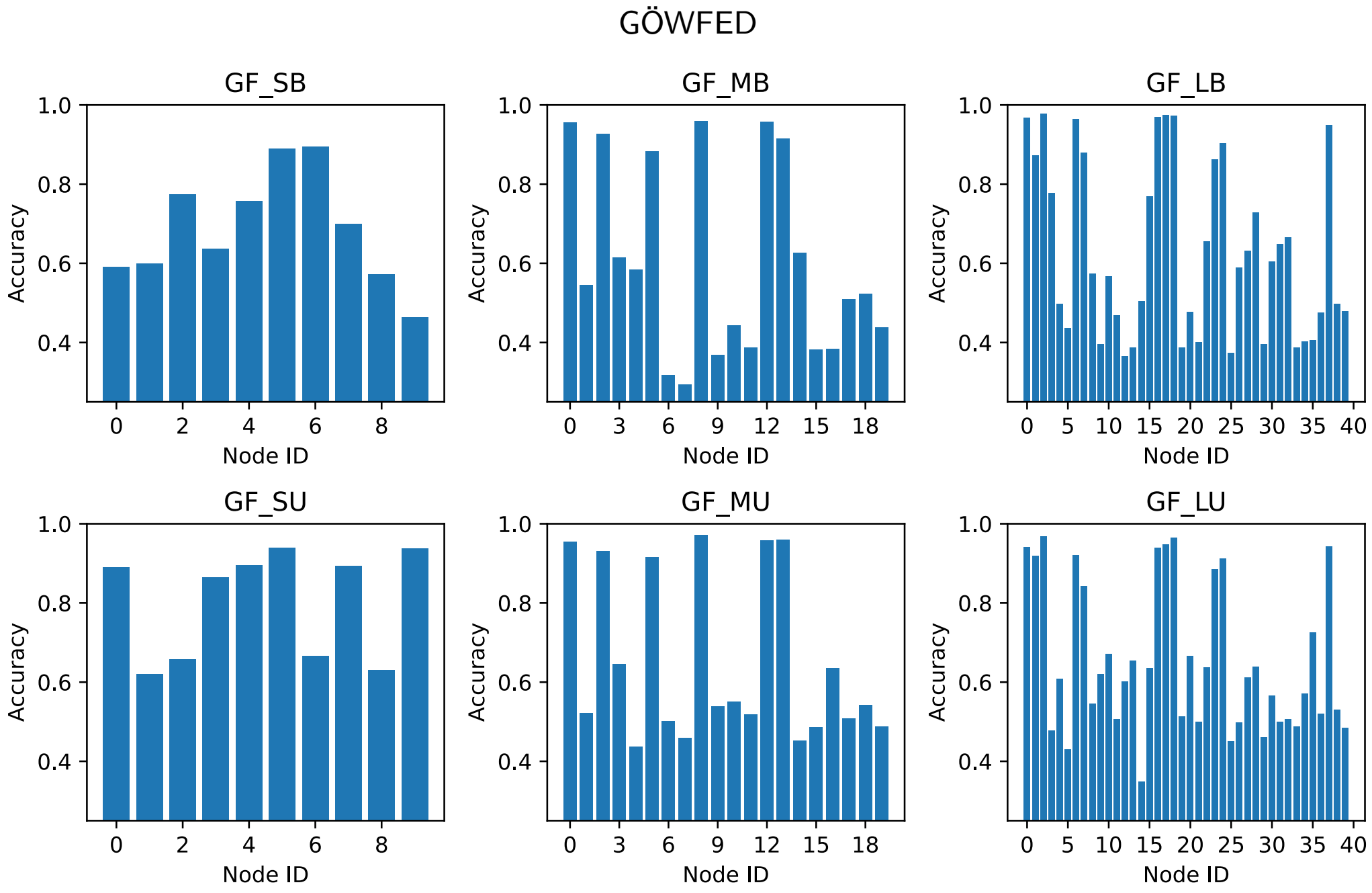

(a) Accuracy per node ID.

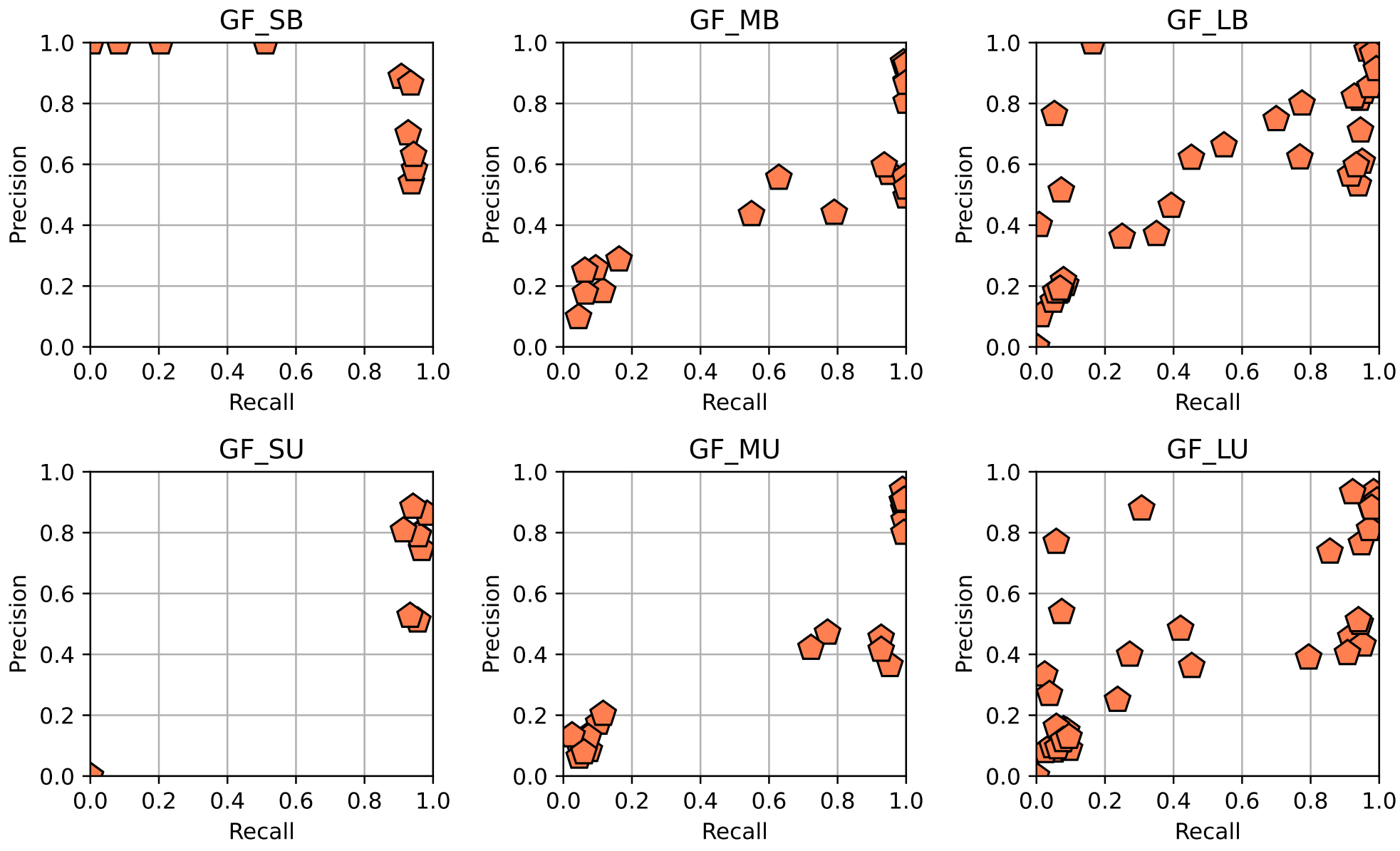

(b) Position in the PR space.

**Figure 10:** Results of learned **GF AM 0.2** models in the test partition of each agent per experiment.

Finally, AM is a custom feature of GF that can not be trivially adapted into GC. Thus, the usage of FL gives a considerable advantage in terms of robustness against poisoning attacks and dummy agents. However, as each party sends its parameters to the server, a party evaluation/challenge mechanism should be added, to avoid tricking the aggregator – specially when working with real IoT devices.

## 7. Conclusions and Future work

GöwFed intends to be an intermediate FL-IDS prototype focused on FL research previous to its deployment in real IoT devices. On the one hand, the modularity of the implementation makes the experimentation with different configurations easier and, indeed, development can be conducted incrementally and debugging can be performed trivially. On the other hand, little adaptations might be necessary to work in real streaming data scenarios, where a single Gower row will be computed instead of the whole matrix. However, the development is partially adapted with two newly implemented functions: `gower_matrix_limit_cols` and `sliced_gower_matrix_limit_cols`. GöwFed is created pursuing the continuity of new experiments using Gower Distance

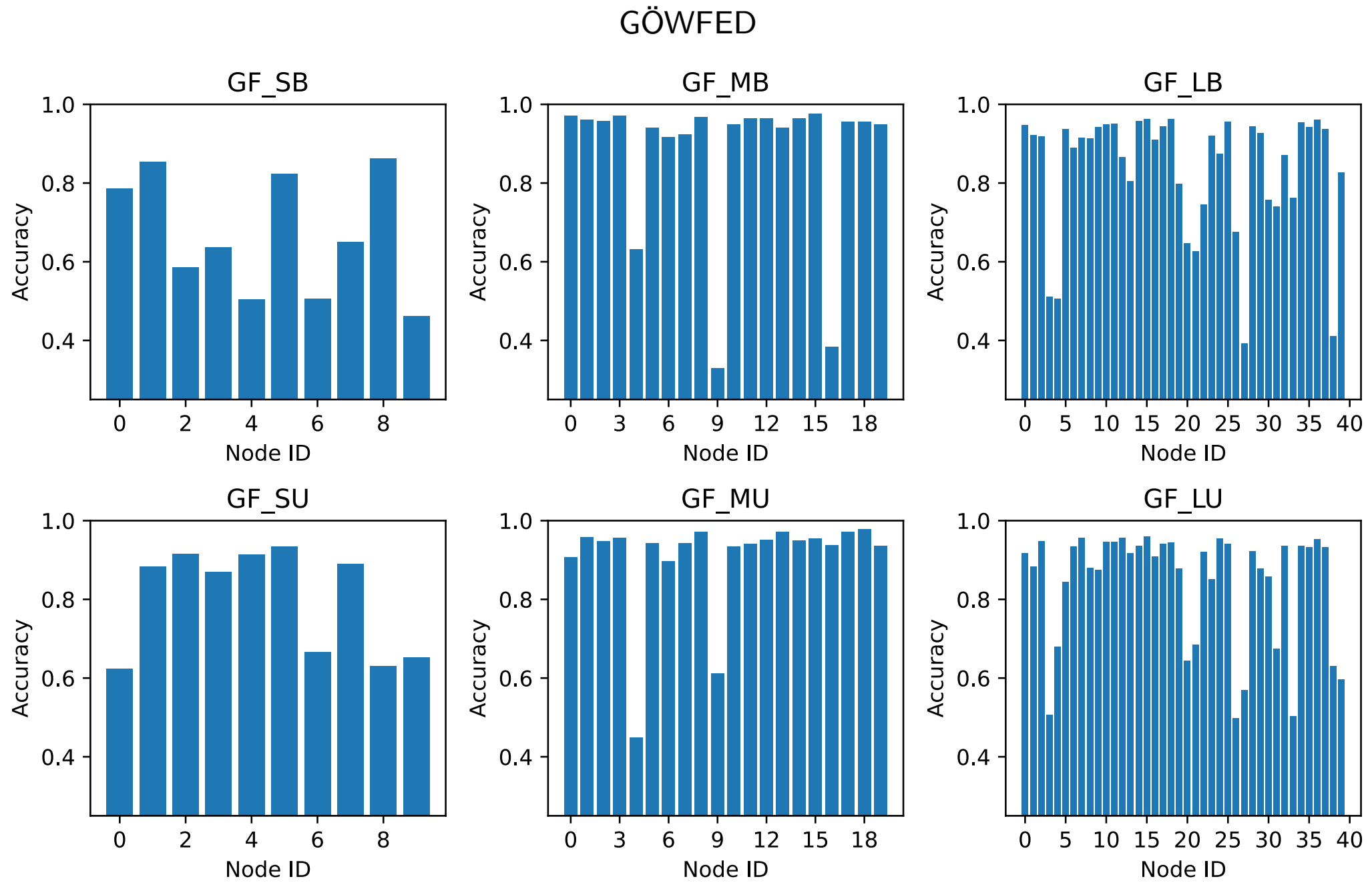

(a) Accuracy per node ID.

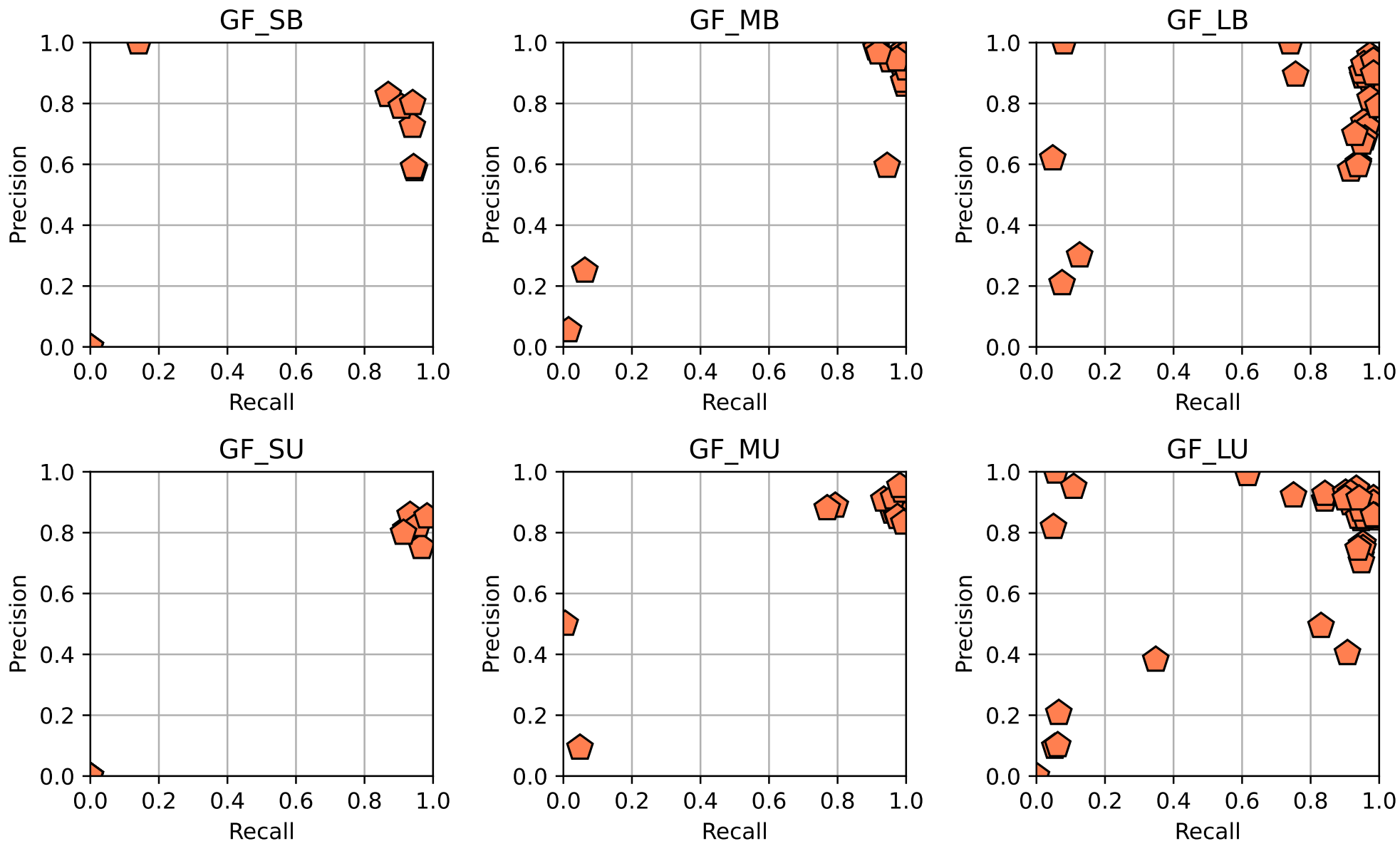

(b) Position in the PR space.

**Figure 11:** Results of learned **GF AM 0.8** models in the test partition of each agent per experiment.

matrix and advanced Deep Learning architectures into a Federated Learning framework.

Regarding the achieved results, GC systems perform slightly better than GF systems in all the explored experiments. As it was expected, GF systems add an extra layer of complexity that somewhat burdens overall performance. However, the comparison is not completely fair due to the scalability issues suffered by the GC versions: As explained in Section 6.4, the number of instances used to create GC and GF system matrices are different due to hardware scalability limitations – e.g., counting with a total number of 120000 instances for federated and 40000 for centralized versions. In addition, epochs (CNL) and communication rounds (FL) do not work similarly, and therefore, model convergence rates can not be directly compared fairly. At this point, it can be concluded that the usage of independent IoT devices under the GöwFed approach, makes the system more scalable than its analogous GC development.

The GF AM version, using 0.8 of best performing agents, achieves good results and makes the system more robust against poisoning attacks. Nonetheless, its potential does not shine in our experiments due to the IID data splitting

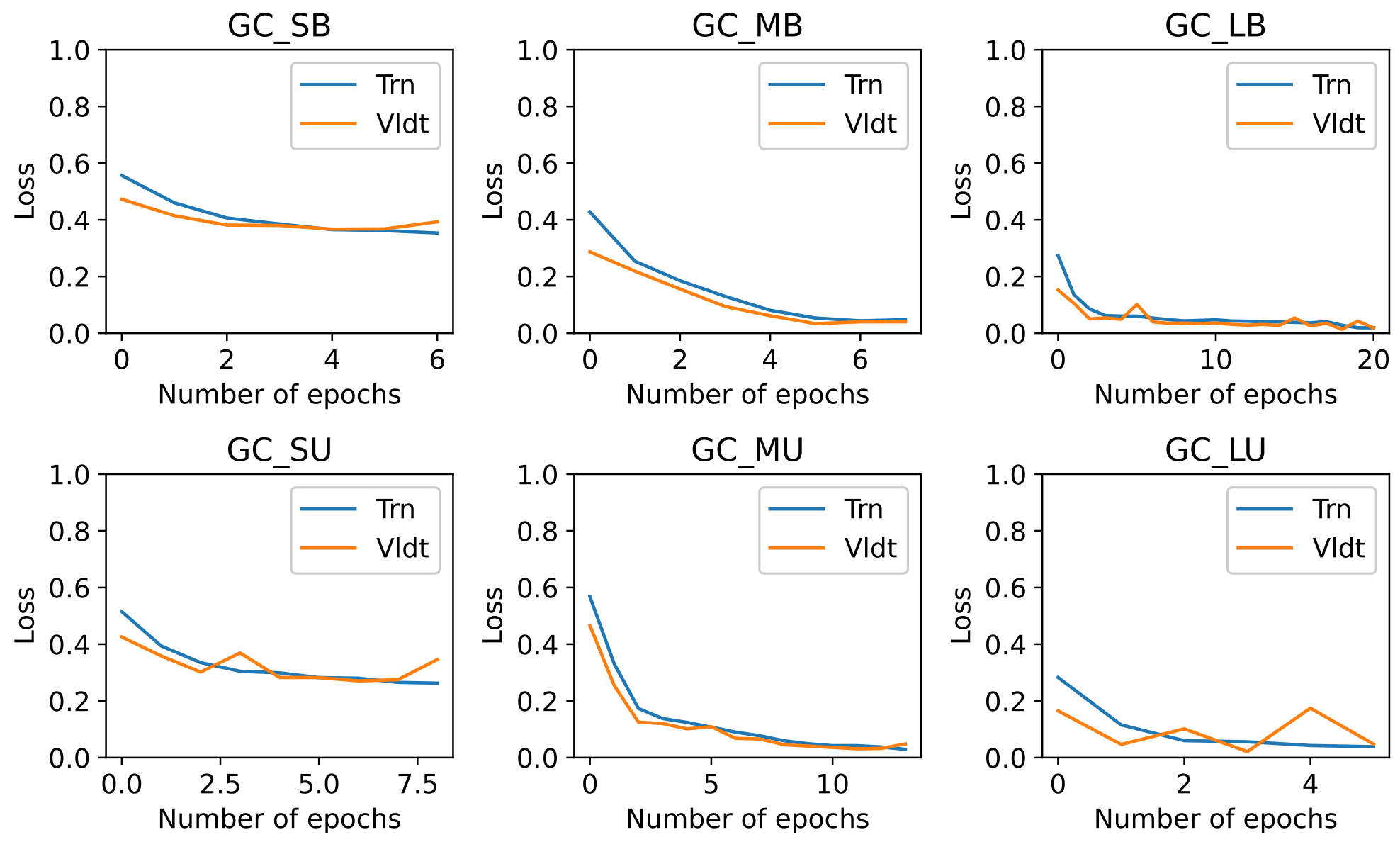

(a) 100 Epochs of training each GC model – with early stopping.

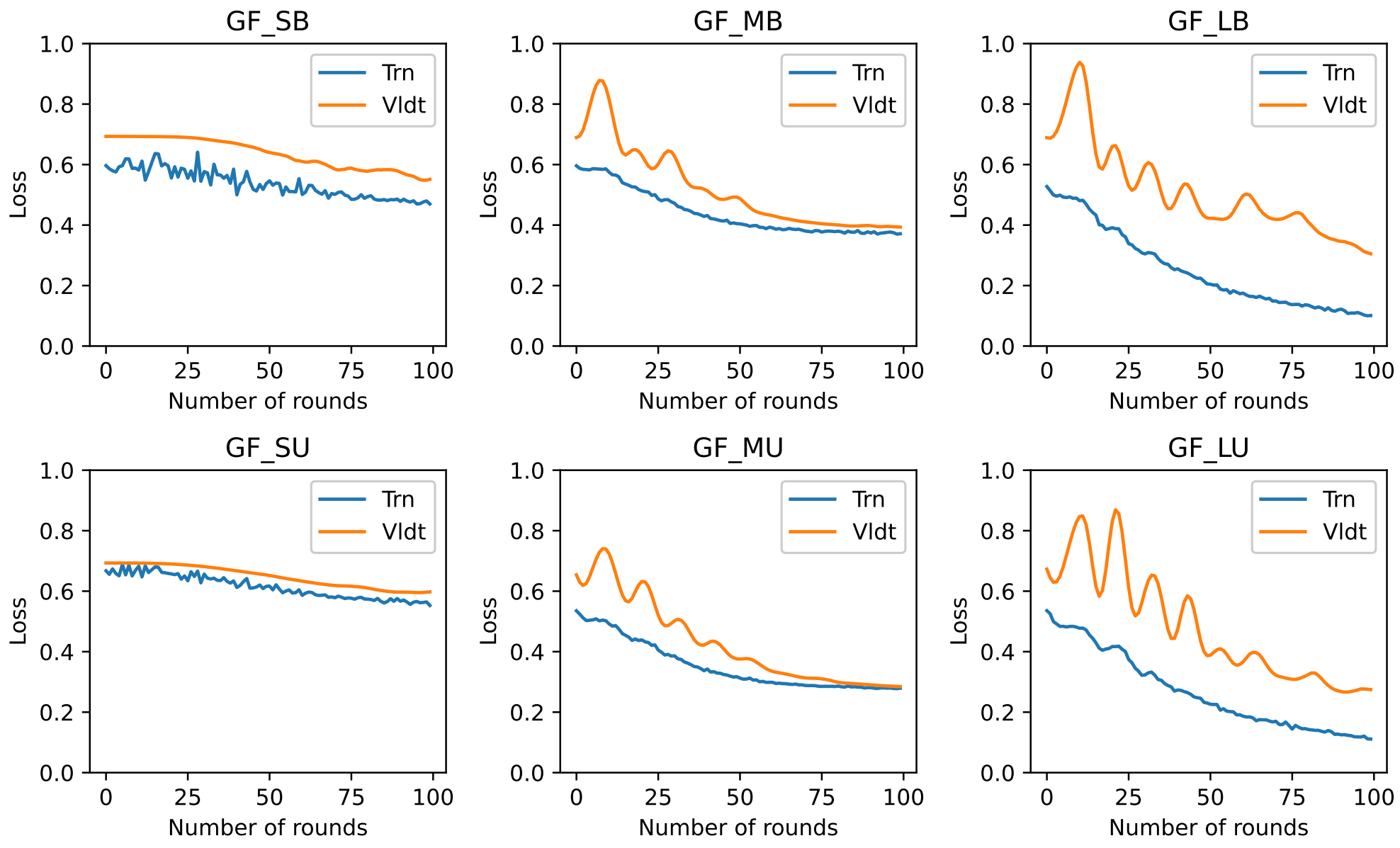

(b) 100 Averaging rounds of each GF global model.

**Figure 12:** Training and validation losses of **vanilla GC and GF** systems respectively in each experiment.

performed among agents – the results show a general worsening compared to the non-AM version. Additionally, non-AM versions count with a weight mechanism that gives (during the averaging process) more importance to nodes with a higher number of instances. As a consequence, results of non-AM version are boosted from the beginning. On the other hand, it is clear that a small percentage of best performing nodes (0.2) is not enough to learn a model that generalizes well – achieving bad results in the majority of the agents. Nevertheless, AM is incorporated to GowFed as part of working with independent model results and parameters; that can not be easily adapted to GC systems – making them more vulnerable. In short, AM is a promising approach that needs to be studied in more detail – testing over heterogeneous devices with non-IID data splitting to make it shine.

After summarizing the results, it can be concluded that the usage of Gower Distance matrices to create a FL-IDS is feasible and doable. That is seconded by the comparison between the losses of GöwFed experiments (see Fig. 12b) and the ones obtained by the toy (non Gower) classifier (in Fig. 5); where GöwFed's loss decreases drastically – especially in experiments FL_LB and FL_LU. In the same way, GF versions perform similarly to their analogous GC ones, in terms of performance and capabilities, so they succesfully achieved the main goal of creating a functional FL-IDS capable of detecting network threats.

In the future, GöwFed will be extended with an incremental learning version tested over data streaming scenarios. This will improve upon the current batch learning version which simulates an artificial environment working over the TON_IOT dataset. However, modifying the implementation to work with real IoT devices should require just some small adaptations; i.e., making the nodes able to process captured datagrams, update local matrices and so on. Moreover, a mechanism to update the matrices in a controlled way should be incorporated as well – avoiding possible outbursts in dimensions growth. In addition, the computation of Gower Dissimilarity among categorical features should be refined to make it smoother than current binary matches of 0 and 1 – making the process more similar to the distances among numerical features.

Similarly, combining our GF system with input data of different nature such as, for instance, time series needs to be carried out – many state-of-the-art approaches work with temporal relationships and RNN. In the same way, a fully distributed FL-IDS has not been developed yet, dispensing with the need of a central orchestration server like the one in FedAVG. The path opened by P. Vanhaesebrouck et al.[45] could be followed to accomplish that task. Nevertheless, novel approaches can trigger new convergence challenges to the nodes. In order to face those incoming challenges, the incorporation of information pre-sharing mechanisms could prevent divergence of the matrices (and models) without compromising the privacy of the agents. Tian et al.[44] propose a mechanism to share a small percentage of local instances among nodes, committing to the mentioned privacy constraint. With respect to model convergence, complementary methods that use data augmentation SMOTE[9] could be used to accelerate the convergence rates of the models[47]. Finally, the incorporation of advanced novelty detection mechanisms could make GöwFed capable of learning unknown threats by clustering the instances of the dissimilarity matrix – going from a binary to a multiclass problem; where anomalies of different nature will be detected.

## Acknowledgments

This work is supported by the Basque Government (projects ELKARTEK21/89 and IT1504-22) and by the Spanish Ministry of Economy and Competitiveness MINECO (PID2019-104966GB-I00). Dr. Javier Navaridas is a Ramón y Cajal fellow from the Spanish Ministry of Science, Innovation and Universities (RYC2018-024829-I).

# Appendix

## GC system configuration

- **Run name:** Name of the current experiment.
- **Training dataset size:** Number of training instances.
- **Test dataset size:** Number of test instances.
- **Balance dataset:** In the GC matrix creation module; balance data to have 50% of normal and 50% of anomalous instances. The new total number of instances will be the double of the class with less appearances.
- **Epochs:** Number of training epochs.
- **Learning rate:** Hyperparameter to specify model learning speed.
- **Batch size:** Hyperparameter.
- **Seed:** Added for replicability.

## GF system configuration

- **Run name:** Name of the current experiment.
- **Node number:** Total number of agents in the network.
- **Training dataset size:** Total number of training instances – summation of all agents' training datasets.
- **Test dataset size:** Total number of test instances – summation of all agents test datasets.
- **Balance dataset:** In the GF matrices creation module; balance data to have 50% of normal and 50% of anomalous instances. The new total number of instances will be the double of the class with less appearances.
- **Total rounds:** Total number of communication rounds – averaging rounds.
- **Nodes per round:** Number of agents taking part in each averaging round.
- **Local epochs per round:** Number of training epochs in each device between averaging rounds.
- **Server learning rate:** Hyperparameter to specify global model learning speed.
- **Client learning rate:** Hyperparameter to specify local models learning speeds.
- **Training batch size:** Same hyperparameter for all local models.
- **Test batch size:** Same hyperparameter for all local models.
- **Seed:** Added for replicability.